\documentclass[10pt,twocolumn,letterpaper]{article}

\usepackage[pagenumbers]{cvpr} % To force page numbers, e.g. for an arXiv version

\definecolor{cvprblue}{rgb}{0.21,0.49,0.74}
\usepackage[pagebackref,breaklinks,colorlinks,allcolors=cvprblue]{hyperref}
\usepackage{multirow}
\usepackage{tablefootnote}

\usepackage{pifont}% http://ctan.org/pkg/pifont
\newcommand{\cmark}{\ding{51}}%
\newcommand{\xmark}{\ding{55}}%

\def\paperID{14460} % *** Enter the Paper ID here
\def\confName{ICCV}
\def\confYear{2025}

\title{Easy3D: A Simple Yet Effective Method for 3D Interactive Segmentation}

\author{Andrea Simonelli \qquad Norman M{\"u}ller \qquad Peter Kontschieder \\ Meta Reality Labs Z{\"u}rich}

\begin{document}
\maketitle
\begin{abstract}
The increasing availability of digital 3D environments, whether through image-based 3D reconstruction, generation, or scans obtained by robots, is driving innovation across various applications. These come with a significant demand for 3D interaction, such as 3D Interactive Segmentation, which is useful for tasks like object selection and manipulation. Additionally, there is a persistent need for solutions that are efficient, precise, and performing well across diverse settings, particularly in unseen environments and with unfamiliar objects. In this work, we introduce a 3D interactive segmentation method that consistently surpasses previous state-of-the-art techniques on both in-domain and out-of-domain datasets. Our simple approach integrates a voxel-based sparse encoder with a lightweight transformer-based decoder that implements implicit click fusion, achieving superior performance and maximizing efficiency. Our method demonstrates substantial improvements on benchmark datasets, including ScanNet~\cite{dai2017scannet}, ScanNet++~\cite{yeshwanth2023scannet++}, S3DIS~\cite{armeni20163d}, and KITTI-360~\cite{liao2022kitti}, and also on unseen geometric distributions such as the ones obtained by Gaussian Splatting~\cite{kerbl20233d}. The project web-page is available here: \url{https://simonelli-andrea.github.io/easy3d} \end{abstract}
\section{Introduction}
\label{sec:intro}

Technological progress has increased the demand for applications that enhance understanding and interaction in digital 3D environments. These environments are in fact now easily obtainable through scene reconstruction (\eg, NeRFs~\cite{mildenhall2021nerf}, Gaussian Splatting~\cite{kerbl20233d}), scene generation, acquisition from laser scans (\eg, LiDAR) or sensors mounted on robots. Among the many tasks that enhance understanding and interaction with 3D environments, 3D Instance Segmentation is essential for enabling applications that involve object selection and manipulation.

\begin{figure}
  \centering
  \begin{subfigure}{0.99\linewidth}
    \includegraphics[width=\textwidth]{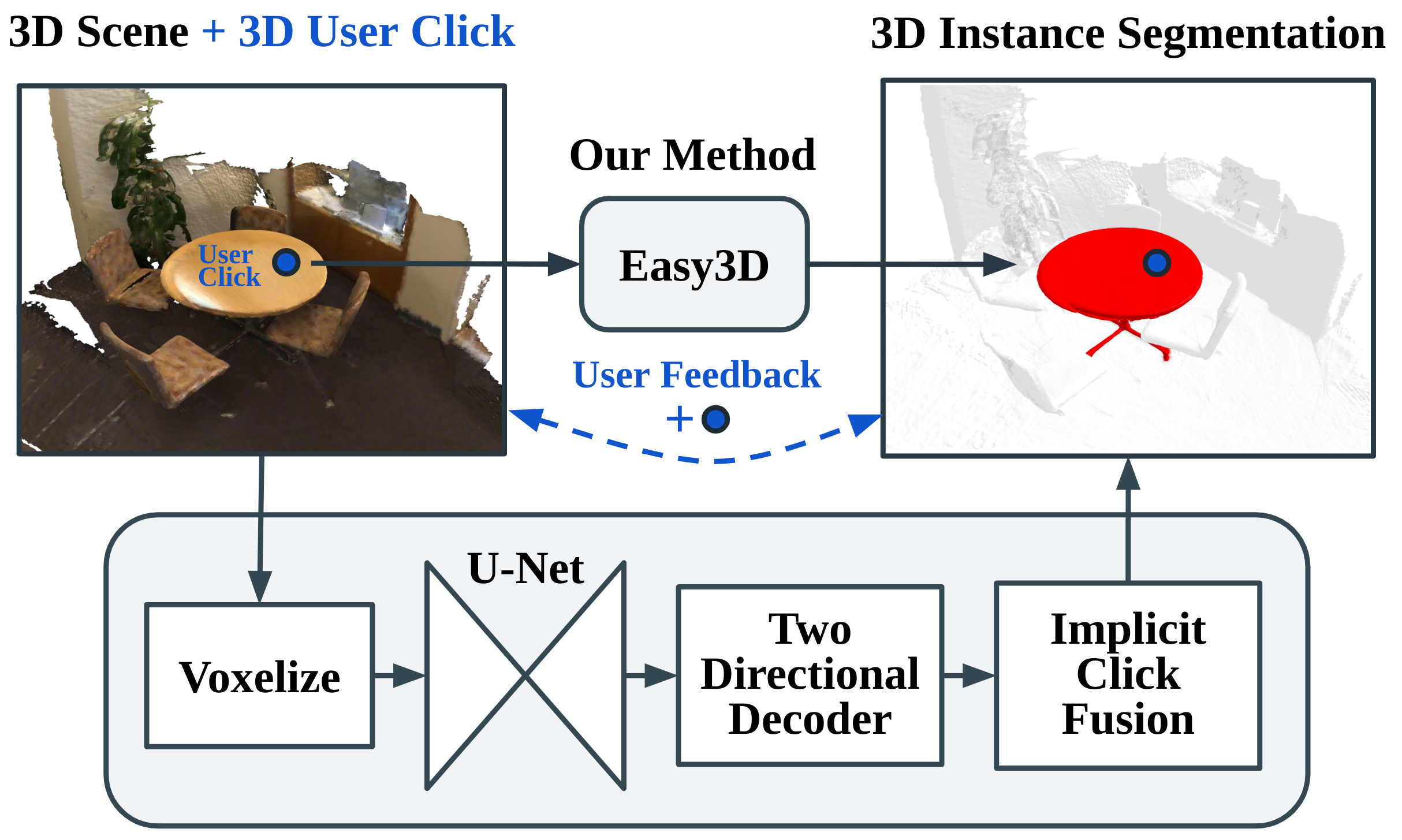}
  \end{subfigure}
  \caption{Example of a typical use-case of Interactive 3D Instance Segmentation (top) and overview of our method's components (bottom). A user interacts with a 3D scene (top left) and defines a 3D click (blue circle) to select an object (table). The scene and clicks are fed to \textbf{Easy3D}, which generates the corresponding instance segmentation mask (top right). After visually inspecting the result (red mask), the user can provide \textit{feedback} and refine the output with additional 3D clicks. To obtain the mask, Easy3D initially maps the scene and user click(s) to a voxelized 3D representation, to then extract scene features using a sparse U-Net. The processed scene and clicks are then fed to a two-directional transformer decoder, which exchanges information through attention to update them. Finally, an implicit click fusion operation is used to predict the output instance segmentation mask. }
  \label{fig:teaser}
  \vspace{-0.25cm}
\end{figure}

While many approaches have been explored in literature \cite{hou20193d, yi2019gspn, engelmann20203d, jiang2020pointgroup, vu2022softgroup, schult2023mask3d, kolodiazhnyi2024oneformer3d, wu2024point, takmaz2023openmask3d, huang2025segment3d}, most are \textit{passive}, offering a fixed set of outputs that may not fully meet user needs. \textit{Interactive} methods, however, enable users to interact with a single 3D click to generate an initial segmentation mask. The result can be immediately observed and further refined by adding positive or negative clicks on the obtained masks, hence allowing for direct user interaction and feedback (as in Fig.~\ref{fig:teaser}). 

Current state-of-the-art architectures typically consist of an encoder and a transformer-based decoder, and can be broadly categorized based on their choice of encoder and click fusion strategy. The click fusion strategy is the method by which information from multiple user clicks, which can be positive or negative, is combined to predict a single object mask. Some methods rely on explicit click fusion, which struggles with out-of-domain and unseen objects (e.g.,~\cite{yue2023agile3d}). Conversely, existing methods that use implicit click fusion (e.g.,~\cite{zhou2024point}) often have inconsistent performance across datasets and rely on architectures (like ViT~\cite{dosovitskiy2020image}) that require a much higher number of parameters.

As applications involving 3D understanding and interaction become more prevalent, it is desirable to have a solution that is simple, efficient, precise, but also capable of performing consistently well across diverse settings, including unfamiliar environments and objects. For these reasons we introduce \textbf{Easy3D}, which combines the strengths of all state-of-the-art approaches by utilizing a voxel-based encoder alongside implicit click fusion. This combination allows our method to achieve superior efficiency and generalization across datasets. Additionally, we incorporate a negative mask component, traditionally used in explicit click fusion, to help the model better understand background elements of the scene. We are the first to demonstrate the positive impact of integrating this component with implicit click fusion, further enhancing the method's performance. By addressing the limitations of existing methods, our method offers an easy but yet robust solution for 3D instance segmentation, paving the way for more effective and adaptable applications in diverse environments.

In summary, the contributions of our paper include:
\begin{itemize}
    \item A simple and efficient voxel-based, interactive 3D instance segmentation method, outperforming existing state-of-the-art methods on both seen and unseen domains, and across challenging benchmark datasets  like ScanNet~\cite{dai2017scannet}, ScanNet++~\cite{yeshwanth2023scannet++}, S3DIS~\cite{armeni20163d}, and KITTI-360~\cite{liao2022kitti}.
    \item The integration of a learned negative embedding in implicit click fusion, improving generalization and segmentation performance.
    \item A targeted analysis on the impact of implicit and explicit click fusion in the context of 3D instance segmentation, providing insights \wrt performance and generalization. 
\end{itemize}

\section{Related Works}
\label{sec:related}

Interactive methods are developed to correct outputs generated by a given system or model, typically through actively provided user feedback. The concept of interactivity has been extensively studied in the 2D setting, starting from pioneering works such as \cite{kass1988snakes, mortensen1995intelligent} to more recent approaches \cite{xu2016deep, li2018interactive, mahadevan2018iteratively, kirillov2023segment, ravi2024sam}, which propose image segmentation methods allowing user feedback via 2D clicks. Among these, Segment Anything~\cite{kirillov2023segment} is the most widely known and adopted method, thanks to its versatile architecture that enables segmentation of any object in an image.

\begin{figure*}
  \centering
  \begin{subfigure}{0.99\linewidth}
    \includegraphics[width=0.99\textwidth]{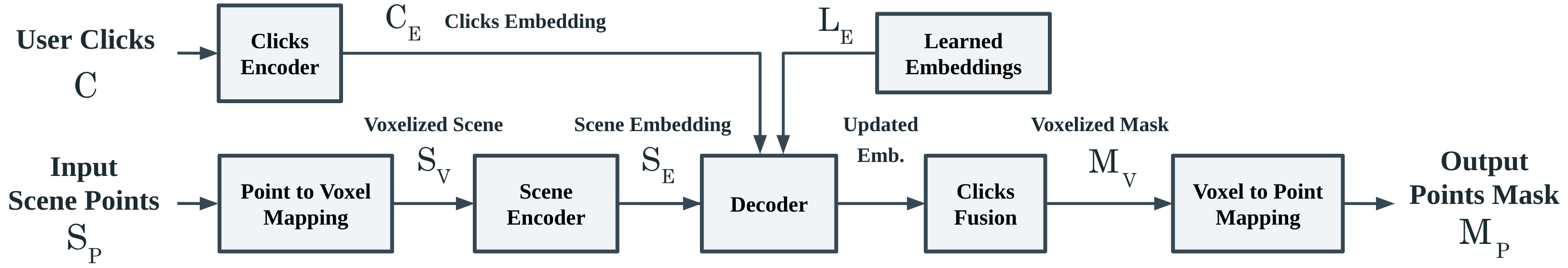}
  \end{subfigure}
  \caption{Architecture of our method for which we provide details in Sec.~\ref{sec:method}. The set of user clicks $C$ (top left) is encoded into the clicks embedding $C_E$. The input scene points $S_P$ (bottom left) are mapped into a voxelized scene $S_V$ and encoded into a scene embedding $S_E$. The clicks and scene embedding, with the additional learned embeddings $L_E$ (e.g. output embeddings), are then fed to the decoder which uses attention operations to update them. The updated embeddings are then fused using a click fusion strategy to obtain the segmentation mask on voxels $M_V$, which is finally mapped back to the original scene points to obtain the output points mask $M_P$.} 
  \label{fig:arch}
\end{figure*} 

In the 3D setting, the majority of the literature focuses on \textit{non-interactive} methods. Early works such as \cite{hou20193d, yi2019gspn, engelmann20203d, jiang2020pointgroup, vu2022softgroup} have recently evolved towards \textit{transformer-based} architectures that leverage attention mechanisms~\cite{vaswani2017attention}, including Mask3D~\cite{schult2023mask3d}, OneFormer3D~\cite{kolodiazhnyi2024oneformer3d}, and PTv3~\cite{wu2024point}. Additionally, works like OpenMask3D~\cite{takmaz2023openmask3d} and Segment3D~\cite{huang2025segment3d} focus on \textit{open-set} and \textit{label-free} 3D segmentation, respectively.
While many approaches exist for Interactive \textit{Semantic} 3D Segmentation~\cite{shen2020interactive, zhi2022ilabel, kontogianni2023interactive, valentin2015semanticpaint}, the task of Interactive 3D \textit{Instance} Segmentation has been more recently introduced by InterObject3D~\cite{kontogianni2023interactive}, aimed at accelerating 3D point cloud annotation. Following this, AGILE3D~\cite{yue2023agile3d} and Point-SAM~\cite{zhou2024point} have further advanced the field, making them the two approaches \textit{most related to ours}. These methods share a similar encoder-decoder architecture and a click fusion strategy, which significantly influences their performance, generalization, and efficiency.

AGILE3D~\cite{yue2023agile3d} employs a voxel-based U-Net~\cite{graham20183d} encoder and a transformer-based decoder derived from Mask3D~\cite{schult2023mask3d}. It utilizes an \textit{explicit} click fusion strategy, a parameter-free \textit{max operation} between the \textit{decoded single-click} masks. Although the method allows users to specify if a click is \textit{positive} or \textit{negative}, indicating whether a particular 3D position should be \textit{included} or \textit{excluded} from the final mask, this information is not utilized inside the decoder. AGILE3D~\cite{yue2023agile3d} also introduces learned negative embeddings, acting as \textit{negative} and position-less clicks to better understand the \textit{non-object} parts of the scene.

Point-SAM~\cite{zhou2024point} instead proposed a 3D version of SAM~\cite{kirillov2023segment}. It uses an encoder which is not voxel-based but uses point-groups, which is an improvement over the traditional k-Nearest Neighbors clustering. It must be noted that, as opposed to voxels, point groups do not have a universal and explicit definition, limiting the method's interpretability and generalization capabilities (as will be demonstrated in the quantitative evaluation in Sec.~\ref{res:interactive}). Point-SAM also uses a large Vision Transformer~\cite{dosovitskiy2020image} (ViT) in their encoder, making the architecture less efficient and considerably increasing the number of parameters. As a decoder, it does not rely on the one from AGILE3D~\cite{yue2023agile3d} but uses the same one as SAM~\cite{zhou2024point}. To fuse the information from multiple clicks, it relies on an \textit{implicit click fusion} strategy. A detailed description of the click fusion strategies will be provided in Sec.~\ref{sec:fusion}. 
%While implemented with a different order of attention operations, the decoders from AGILE3D~\cite{yue2023agile3d} and Point-SAM~\cite{zhou2024point} are found to perform equally.

Our method builds upon these approaches by combining the strengths of voxel-based encoders with implicit click fusion, addressing the limitations of existing methods in terms of efficiency and generalization across diverse datasets.
\section{Method}

In this section we first introduce the task of Interactive 3D Instance Segmentation (Sec.~\ref{subsec:task}), to then delve into the details of the method (Sec.~\ref{sec:method}), to finally detail the training protocol and losses (Sec~\ref{sec:training_protocol}). 

\vspace{-0.30cm}
\subsection{Interactive 3D Instance Segmentation}
\label{subsec:task}
In the most common scenario, an Interactive 3D Instance Segmentation pipeline follows the one in Fig.~\ref{fig:teaser}. A user starts by defining a first (positive) click on the target object (\eg, the table), to then observe the corresponding, generated output mask. If the segmentation result is unsatisfactory, the user can provide additional feedback through more 3D clicks (one or more, each either positive or negative), to then obtain an updated output mask. The feedback process can be repeated until the user is satisfied with the result, ideally obtained from a minimal number of interactions.

More formally, the input to the method is mainly defined by two elements, a \textit{3D scene} and a set of one or more \textit{3D user clicks}. The 3D scene is the digital 3D environment where the user wants to segment an object, \eg~a colored point cloud obtained via a 3D reconstruction method. We refer to the input 3D scene as $S_P \in \mathbb{R}^{\mathsf{N_P} \times \mathsf{N_{F_P}}}$, where $\mathsf{N_P}$ is the number of scene points and $\mathsf{N_{F_P}}$ is the number of features related to each point. In all our experiments $\mathsf{N_{F_P}}=6$, which is due to the fact that each point is represented by its 3D position and color. 

The set of \textit{3D user clicks} is instead what allows the method to know which object should be segmented inside the 3D scene. We will refer to it as $C =\{c_1, c_2, ..., c_\mathsf{N_C}\} \in \mathbb{R}^{\mathsf{N_C} \times {\mathsf{N_{F_C}}}}$, where $\mathsf{N_C}$ is the number of clicks and $\mathsf{N_{F_C}}$ is the number of click features. In our method $\mathsf{N_{F_C}}=4$ due to the fact that each click $c_i$ is represented by its 3D coordinates and \textit{label}. The click label is a specific feature of interactive methods, as it usually represents the type of click feedback that will be provided to the model. In the methodology introduced by SAM~\cite{kirillov2023segment}, the user can define \textit{positive} or \textit{negative} clicks depending if a particular position should be included (positive) or excluded (negative) in the output.

Given the input scene $S_P$ and the set of input user clicks $C$, the Interactive 3D Segmentation method has to predict an output segmentation mask $M_P \in \mathbb{R}^{\mathsf{N_P}}$, which denotes the set of scene points that belong to the object described by the user clicks. 

\subsection{Method Details}
\label{sec:method}
We will now go into the details of the method, for which we provide a high-level overview in Fig.~\ref{fig:arch}. We will start by explaining the scene preprocessing (Sec.~\ref{sec:scene_preprocessing}) and scene encoding (Sec.~\ref{sec:scene_encoding}), followed by the clicks encoding (Sec.~\ref{sec:click_encoding}), to then describe the decoding (Sec.~\ref{sec:decoder}) and click fusion (Sec.\ref{sec:fusion}), describe the notion of learned negative embedding and how it is used in ours and other methods (Sec.~\ref{sec:negative_embeddings}), to finally explain the final output postprocessing (Sec.~\ref{sec:postprocessing}) steps. 

\vspace{-0.2cm}
\subsubsection{Scene Preprocessing}
\label{sec:scene_preprocessing}
The scene preprocessing involves the mapping of the input scene pointcloud $S_P\in\mathbb{R}^{\mathsf{N_P} \times \mathsf{N_{F_P}}}$ into its voxelized version $S_V\in\mathbb{R}^{\mathsf{N_V} \times \mathsf{N_{F_V}}}$, where $\mathsf{N_V}$ is the number of voxels. Similarly to points, each voxel is defined by its 3D coordinates and color, therefore $\mathsf{N_{F_P}} = \mathsf{N_{F_V}}$. The voxel coordinates are obtained with a discretization of the pointcloud coordinates with a predefined voxel metric resolution $\mathsf{V_S}$, and the color of each voxel is computed by averaging the color of all the points that fall into it. 
Note that the just described voxelization is in stark contrast with the process done in Point-SAM~\cite{zhou2024point}, which instead uses a tokenizer to group points and create group tokens. The groups are created using a predefined number of groups and points per group, with a novel method that improves over traditional k-Nearest Neighbors (kNN). For the details about this procedure, we invite to refer directly to~\cite{zhou2024point}.
The voxelization offers a number of advantages over the point groups: 1) it reduces the dimensionality of the problem while retaining an explicit, universal metric voxel resolution (\ie $\mathsf{V_S}$); 2) it allows us to rely on established and efficient sparse libraries for optimized computation (\ie sparse convolutions); 3) it is a domain-independent representation which lets the model be more robust \wrt changes in object type, size, as well as scene geometry and point density, achieving better generalization as proven by our quantitative results in both seen and unseen benchmarks (Sec.~\ref{res:interactive}). 

\vspace{-0.2cm}
\subsubsection{Scene Encoder}
\label{sec:scene_encoding}
The scene encoder takes as input the scene voxelization $S_V\in\mathbb{R}^{\mathsf{N_V} \times \mathsf{N_{F_V}}}$ and outputs the scene embedding $S_E$. 
We encode the scene using a U-Net~\cite{graham20183d}, which uses a series of sparse convolutions followed by a linear layer to go from the voxelized scene representation to the output-encoded scene representation $S_E \in \mathbb{R}^{\mathsf{N_V} \times \mathsf{N_{F_E}}}$, where $\mathsf{N_{F_E}}$ defines the number of features of the embedding. Differently from our method, Point-SAM~\cite{zhou2024point} uses a ViT encoder (large), which takes as input the point groups tokens and returns the corresponding point groups token embeddings. In order to provide some clarity and parallelism, we can think about a voxel as a point group, which includes points that belong to a specific region of the 3D space determined by the voxel resolution. On the other hand, Point-SAM's groups can include points that are in an arbitrarily big and skewed region, which is defined by the input pointcloud geometry and density, as well as by the number of groups and points per group predefined by hyperparameters. This should help to further highlight the fact that, if trained on a specific dataset with a defined geometry and density, a model based on point-groups will likely struggle to generalize towards data that does not match the same properties. 

\vspace{-0.2cm}
\subsubsection{User Clicks Encoder}
\label{sec:click_encoding}
The clicks encoder takes as input the set of input user clicks $C \in \mathbb{R}^{\mathsf{N_C} \times {\mathsf{N_{F_C}}}}$ and outputs the clicks embedding $C_E \in \mathbb{R}^{\mathsf{N_C} \times {\mathsf{N_{F_E}}}}$. In our method and Point-SAM~\cite{zhou2024point}, we follow SAM~\cite{kirillov2023segment} and encode the user clicks as a combination of \textit{positional} and \textit{label} encoding. The positional encoding follows \cite{kirillov2023segment, tancik2020fourier}, and has the purpose of informing the model about the 3D position of each click with respect to the 3D scene's extent. The label encoding informs the model if a particular position in the scene should be included (\textit{positive} click) or excluded (\textit{negative} click) in the final output mask. The \textit{positional} and \textit{label} encoding have the same dimensionality of $\mathsf{N_{F_E}}$ and, as done in SAM~\cite{kirillov2023segment}, the click embedding is given by their sum. Note that this encoding is different with respect to AGILE3D~\cite{yue2023agile3d}, which does \textit{not} include the positive/negative label but rather encodes the click's position in time (e.g.~1st, 2nd, Nth) to make the model aware of the order in which the clicks were created.

\begin{figure}
  \centering
  \begin{subfigure}{0.99\linewidth}
    \includegraphics[width=0.99\textwidth]{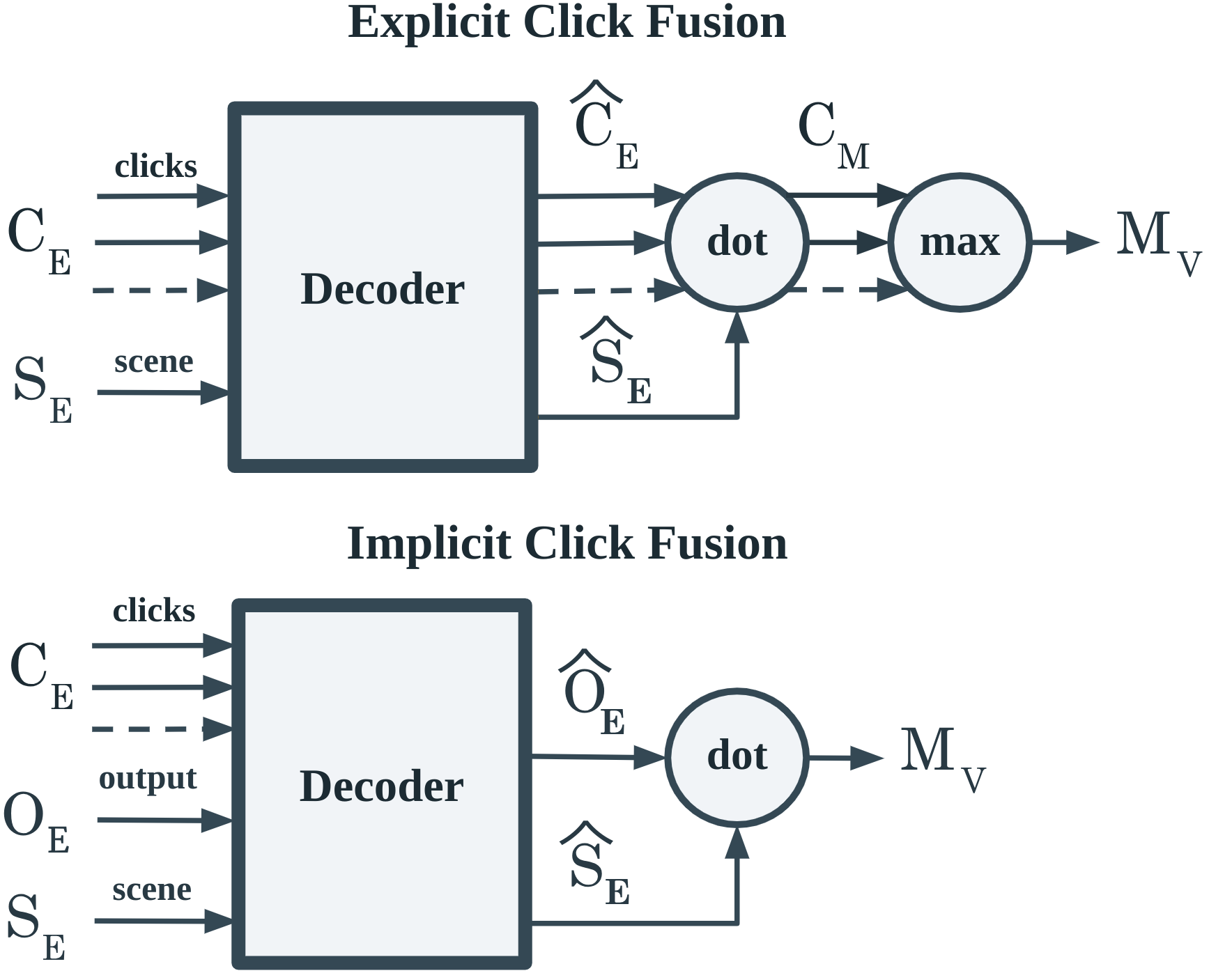}
  \end{subfigure}
\caption{Depiction of explicit click fusion as in AGILE3D~\cite{yue2023agile3d} (top) and implicit click fusion as used in our method (bottom).} 
  \label{fig:decoders}
  \vspace{-0.4cm}
\end{figure}

\vspace{-0.20cm}
\subsubsection{Decoder}
\label{sec:decoder}
As shown in Fig.~\ref{fig:arch}, the decoder takes as input the scene embedding $S_E$ and the clicks embedding $C_E$ to predict their updated version $\widehat{S_E}$ and $\widehat{C_E}$ (for clarity, the learned embeddings $L_E$ will be introduced later). It does so by updating the scene and clicks embeddings using a series of attention layers. Following SAM~\cite{kirillov2023segment}, we use a decoder based on a \textit{Two-Way Transformer}, which performs a combination between self-attention and cross-attention operations. Note that also AGILE3D~\cite{yue2023agile3d} uses a similar decoder with different order of attention operations, which we found do not affect the final performance. What is crucial is the two-directional set of attention operations, which by following the original~\cite{vaswani2017attention} formulation, is achieved by alternating between using the scene and clicks embedding as query and key/value, allowing to update both conditioned on the other and maximizing the flow of information. We invite to refer to the supplementary material or \cite{kirillov2023segment, yue2023agile3d} for details on the attention layers and the precise architecture. 

\subsubsection{Click Fusion and Output Mask Prediction}
\label{sec:fusion}
The click fusion takes as input the updated scene and clicks encoding to obtain a \textit{single} output mask from \textit{one or multiple} clicks. In literature, this operation is performed in an explicit or implicit way, by AGILE3D~\cite{yue2023agile3d} and Point-SAM~\cite{zhou2024point} respectively. We will now go into the details of both approaches, depicted in Fig~\ref{fig:decoders}, and later showing their impact in the experiments section (Sec.~\ref{sec:experiments}).

\textbf{Explicit Click Fusion:} given the updated scene and click embeddings, this fusion performs: 1) Dot product ($\cdot$) between them to obtain the clicks output masks $C_M$, one for each click; 2) Prediction of the fused output mask with a \textit{max operation} among the clicks masks. In case there are both positive and negative clicks, the fused output mask $M_V$ is defined by where the positive clicks masks~$>$~negative clicks masks. It must be noted that, in the explicit click fusion of AGILE3D~\cite{yue2023agile3d}, the decoder's attention operations are \textit{not aware} of the click's label (positive/negative), and the only place where the information is used is in the max operation to determine the output mask.

\textbf{Implicit Click Fusion:} this is the same operation performed by SAM~\cite{kirillov2023segment} in 2D Interactive Instance Segmentation, and involves the introduction of an additional \textit{learned embedding}, i.e. \textit{output embedding} ($O_E$), which is simply concatenated to the clicks embedding in the decoder as an additional click, and participates to the attention operations with the purpose of combining the information of the user-defined clicks. After the decoding, the output mask $M_V$ is obtained with a dot product between the updated scene embedding ($\widehat{S_E}$) and the updated output embedding ($\widehat{O_E}$), without the need of predicting the clicks masks. Note that, differently from the explicit click fusion, the implicit click fusion is aware of the click's label. As the output embedding mask is used as the only output, the attention operations in the decoder need and will learn how to use this information to come up with a meaningful solution.  

\vspace{-0.20cm}
\subsubsection{Learned Negative Embeddings}
\label{sec:negative_embeddings}
In addition to the user-defined clicks, AGILE3D~\cite{yue2023agile3d} introduces a set of learned embeddings, which act as position-less negative clicks. These embeddings are simply concatenated to the list of clicks embeddings in the decoder (similarly to the output embedding of the implicit click fusion), and participate to the explicit max operation as negative clicks. During training, these embeddings will basically learn which parts of the scene are usually part of the background. This is an interesting technique from AGILE3D which, by adding these learned negative embeddings, can help the model to discard the background even with very few user clicks (e.g. one click). 
In this paper, we are the first ones to investigate the use of these learned negative embeddings in the implicit fusion. To do this, we added a second learned output embedding, which we refer as \textit{negative} output embedding, to the procedure explained above. We then obtain its mask with the same dot operation and define the fused output mask $M_V$ as where the positive output embedding mask~$>$~negative output embedding mask. 
As it will be shown later in a targeted ablation study (Sec.~\ref{sec:ablation_click_fusion_neg_emb}), the learned negative embedding consistently improves the segmentation performance also with implicit fusion. 

\vspace{-0.20cm}
\subsubsection{Postprocessing}
\label{sec:postprocessing}
The output mask obtained after the click fusion is still related to the scene voxelization so, in order to get the final output mask, we use the initial scene preprocessing voxelization (Sec.~\ref{sec:scene_preprocessing}) and map $M_V$ back to the original input pointcloud obtaining $M_P\in \mathbb{R}^{\mathsf{N_P}}$. Note that, following AGILE3D~\cite{yue2023agile3d}, this final point-based mask is computed only at inference time. During training and in all the losses computation, we use the voxel-based mask $M_V$.

\vspace{0.10cm}
\subsection{Training Pipeline}
\label{sec:training_protocol}
An interactive method requires a specific training protocol which simulates a user interaction. Next, we provide the details for the protocol used in our work (Sec~\ref{subsubsec:simulation}), as well as the losses (Sec~\ref{subsubsec:losses}). 

\subsubsection{Simulated User Interaction}
\label{subsubsec:simulation}
Due to the fact that providing user feedback during training is unfeasible, we follow SAM~\cite{kirillov2023segment} and AGILE3D~\cite{yue2023agile3d} and \textit{simulate} the user feedback through an automatic click selection procedure. The procedure is as follows: 1) Given the scene and its ground-truth instance annotation labels, we randomly select one object to be segmented; 2) Knowing which parts of the scene belong to that object, we start by simulating the first click $c_1$ on the \textit{center} of the object. Following AGILE3D~\cite{yue2023agile3d}, we define the \textit{center} as the point with the \textit{highest distance} with respect to any non-object point in the scene; 3) We provide this first click $c_1$ to the model and obtain the first segmentation mask; 4) We analyze the errors in the mask (i.e.~False Positives and False Negatives) and, similarly to AGILE3D~\cite{yue2023agile3d}, we select the one at the center of the biggest error region. Note that the selected error can either be a \textit{False Negative}, which will result in a \textit{positive} click, or a \textit{False Positive}, which will result in a \textit{negative} click; 5) We add the selected click ($c_2$) to the set of clicks, now $C=\{c_1$, $c_2\}$, feed the updated set to the model, and so on. These steps are repeated for a predefined number of clicks, which in our experiments is $\mathsf{N_C}=10$. 

\vspace{-0.20cm}
\subsubsection{Losses}
\label{subsubsec:losses}
We follow AGILE3D~\cite{yue2023agile3d} and supervise our method with a combination of DICE~\cite{deng2018learning} and Cross Entropy losses. During the simulated user interaction procedure explained in Sec.~\ref{subsubsec:simulation}, we compute these two losses at each iteration. Once the number of clicks has reached $\mathsf{N_C}$, we sum and backpropagate the total loss just once. We weight each loss equally, normalizing the total loss by the number of clicks. 

\vspace{-0.20cm}
\subsubsection{Implementation Details}
\label{subsubsec:impl_details}
We implemented our method using PyTorch~\cite{paszke2017automatic} and SpConv~\cite{spconv2022}. Similarly to AGILE3D~\cite{yue2023agile3d}, we use attention modules with 8 heads, while the MLP network has a dimensionality of 1024. Without any pretraining, we train our model for 1k epochs, setting a polynomial learning rate that starts at $1e{-4}$ and is then decayed using a polynomial with an exponent of 0.9. More details are provided in the supplementary material. 

\section{Experiments}
\label{sec:experiments}
In this section we first provide details about the datasets (Sec.~\ref{datasets}) and metrics (Sec.~\ref{sec:metrics}), before we explain the evaluation protocol (Sec.~\ref{sec:eval_protocol}), and finally provide quantitative (Sec.~\ref{sec:quantitative_results}) and qualitative results (Sec.~\ref{sec:qualitative_results}). 

\begin{table*}
    \centering
    \resizebox{1.2\columnwidth}{!}{%
    \begin{tabular}{l|l|c|c|c|c|c}
        \toprule
        \textbf{Test Dataset} & \textbf{Method} & \textbf{IoU@1} & \textbf{IoU@2} & \textbf{IoU@3} & \textbf{IoU@5} & \textbf{IoU@10} \\ 
        \toprule
        \multirow{2}{*}{ScanNet40} & AGILE3D~\cite{yue2023agile3d} & 63.0 & 70.6 & 75.1 & \textbf{79.7} & \textbf{83.5} \\ 
                                   & Ours    & \textbf{68.2} & \textbf{74.6} & \textbf{77.3} & 79.6 & 81.7 \\ 
        \hline       
        \multirow{3}{*}{S3DIS}     & AGILE3D~\cite{yue2023agile3d}   & 58.5 & 70.7 & 77.4 & 83.6 & \textbf{88.3} \\ 
                                   & Point-SAM~\cite{zhou2024point} & 38.8 &  n/a   & 67.1 & 72.2 & 80.6 \\
                                   & Ours      & \textbf{65.7} & \textbf{76.0} & \textbf{80.8} & \textbf{84.9} & 87.8 \\
        \hline
        \multirow{3}{*}{KITTI-360} & AGILE3D~\cite{yue2023agile3d}   & 34.8 & 40.7 & 42.7 & 44.4 & 49.6 \\ 
                                   & Point-SAM~\cite{zhou2024point} & 44.0 & n/a    & \textbf{67.1} & 72.2 & 80.8 \\
                                   & Ours      & \textbf{46.3} & \textbf{58.7} & 66.7 & \textbf{76.2} & \textbf{83.6} \\
        \hline
        \multirow{2}{*}{ScanNet++} & AGILE3D~\cite{yue2023agile3d}   & 49.5 & 58.4 & 63.3 & \textbf{68.4} & \textbf{74.2} \\ 
                                   & Ours      & \textbf{54.8} & \textbf{61.4} & \textbf{64.7} & 67.9 & 71.3 \\        
        \bottomrule
    \end{tabular}}
    \caption{Quantitative results on single-object interactive segmentation. All models are trained only on ScanNet40.}
    \label{tab:interactive}
    \vspace{-0.3cm}
\end{table*}

\subsection{Datasets}
\label{datasets}
For our quantitative evaluation we closely follow the datasets used in AGILE3D~\cite{yue2023agile3d} which include ScanNet~\cite{dai2017scannet}, an indoor dataset comprising of $\approx$ 1500 meshes obtained using a laser scan, Stanford Large-Scale 3D Indoor Spaces Dataset (S3DIS)~\cite{armeni20163d}, another indoor dataset captured in the Stanford campus, and KITTI-360~\cite{liao2022kitti} which is an outdoor dataset primarily focused on autonomous driving applications. Similarly to AGILE3D~\cite{yue2023agile3d} we use ScanNet in two versions, one comprising of 40 object categories (ScanNet40) and another with only the subset of 20 Instance Segmentation benchmark classes (ScanNet20)\footnote{More ScanNet details can be found in the \href{https://kaldir.vc.in.tum.de/scannet_benchmark/documentation}{official documentation}.}. In order to further test the generalization capabilities of the methods, we also include ScanNet++~\cite{yeshwanth2023scannet++}, which is a novel indoor dataset that offers high-quality reconstructions and instance annotations on a variety of 100+ object categories. Lastly, to provide a quantitative evaluation on unseen geometric distributions, we propose a Gaussian Splatting version of ScanNet (GS-ScanNet40), obtained by running the state-of-the-art open-source method SplatFacto~\cite{nerfstudio} on the ScanNet scenes. More details about ScanNet++~\cite{yeshwanth2023scannet++} and GS-ScanNet40 are in the supplementary material.  

\subsection{Metrics}
\label{sec:metrics}
We closely follow AGILE3D~\cite{yue2023agile3d} and InterObject3D~\cite{kontogianni2023interactive}, providing results using the IoU@$i$ score, \ie~the Intersection-over-Union between the predicted and ground-truth instance mask after $i$ clicks. We provide results for a maximum of $\mathsf{N_C}=10$ clicks, particularly focusing on results with $\leq3$ clicks which are the most relevant ones for real-time, virtual reality applications as discussed in Sec.~\ref{sec:intro}.

\subsection{Evaluation Protocol}
\label{sec:eval_protocol}
We closely follow AGILE3D~\cite{yue2023agile3d} in their \textit{single-object} setting, where we assume to have a 3D scene where a single object has to be segmented. As done in the training protocol (and by AGILE3D~\cite{yue2023agile3d}) we simulate a user interaction also during evaluation, following the procedure described in Sec.~\ref{subsubsec:simulation}. In brief, we place the first click on the center of the object and the following clicks at the center of the biggest predicted error region, until we reach a total of $\mathsf{N_C}$ clicks. For each click $c_i$ we compute the IoU@i as described in Sec.~\ref{sec:metrics}. For what concerns fair evaluation with the existing literature, we closely follow AGILE3D~\cite{yue2023agile3d} by: 1) training and evaluating using their released preprocessed data; 2) no model pretraining 3) evaluating using their selected instances; 4) compute metrics using their \href{https://github.com/ywyue/AGILE3D/tree/fcb8444ca2e2bd8270d349874b2e083c78e93b7d}{official repository}. Also, we follow AGILE3D~\cite{yue2023agile3d} in setting the voxel resolution $\mathsf{V_S}$=5cm for all our experiments. 

\subsection{Quantitative Evaluation}
\label{sec:quantitative_results}
In this section we provide multiple quantitative evaluations, starting from a direct comparison with interactive methods (Sec.~\ref{res:interactive}), to then compare with non-interactive methods (Sec.~\ref{res:non_interactive}) and finally ablation studies (Sec.~\ref{subsec:ablation}). 

\subsection{Comparison with Interactive Methods}
\label{res:interactive}
In Tab.~\ref{tab:interactive} we report a quantitative comparison with state-of-the-art interactive methods. All methods have been trained only on ScanNet40\footnote{Point-SAM~\cite{zhou2024point} does not currently provide a complete quantitative evaluation of a model trained only on ScanNet40 (like ours). 
%At the time of submission, the paper does not provide a complete, reproducible training code or a ScanNet40 checkpoint for fair evaluation with our method. 
We report all the available results of a model trained only on ScanNet40 from \cite{zhou2024point}.} and evaluated on the dataset shown as \textit{Test Dataset} in the first column. All methods use a voxel size of 5cm, with the exclusion of Point-SAM~\cite{zhou2024point} which does not rely on voxels and does not have a direct, objective way to set an equivalent resolution. Our method is the one that shows the most consistent performance across datasets, with particular improvements on unseen objects and unseen domains (i.e.~S3DIS~\cite{armeni20163d}, KITTI-360~\cite{liao2022kitti}). A relevant thing to note is how SAM~\cite{kirillov2023segment}-based methods like ours (and Point-SAM~\cite{zhou2024point}) achieve considerable better results than AGILE3D~\cite{yue2023agile3d} on KITTI-360~\cite{liao2022kitti}, which represents the most challenging \textit{unseen} domain. As it will be shown in a targeted ablation study (Sec.~\ref{sec:ablation_click_fusion_neg_emb}) this is due to the \textit{implicit fusion} of clicks which, even on completely unseen objects, still allows the model to better derive what the target object is. On the other hand, AGILE3D~\cite{yue2023agile3d} relies on an \textit{explicit click fusion} which forces the model to reason about each click independently. In the case of completely unseen objects, the performance is still driven by the combination of single-click masks. We also note that AGILE3D~\cite{yue2023agile3d} achieves a comparable performance w.r.t. our method when the number of clicks gets high ($\mathsf{N_C}>5$) and the setting is similar to the training one, suggesting the fact that the \textit{explicit click fusion} could be beneficial in such scenarios. Please refer to Sec.~\ref{sec:ablation_click_fusion_neg_emb} for more details. Lastly, we note that our method outperforms Point-SAM on almost all benchmarks and has a much more consistent performance across datasets. This validates our choice of voxels as a core 3D representation and encoder architecture. 

% \vspace{-0.2cm}
\begin{table}
    \centering
    \resizebox{0.9\columnwidth}{!}{%
    \begin{tabular}{l|l|c|c|c}
        \toprule
        \textbf{Setting} & \textbf{Method} & mAP & $AP_{50\%}$  & $AP_{25\%}$ \\
        \toprule
        \multirow{3}{*}{ScanNet20~$\rightarrow$~Scannet20} & Mask3D~\cite{schult2023mask3d}   & 51.5 & 77.0 & 90.2 \\
        & AGILE3D~\cite{yue2023agile3d}  & 53.5 & 75.6 & 91.3 \\
        & Ours     & \textbf{56.1} & \textbf{79.5} & \textbf{93.1} \\
        \midrule
        \multirow{3}{*}{ScanNet20~$\rightarrow$~ScanNet40} & Mask3D~\cite{schult2023mask3d}   & 5.3 & 13.1 & 24.7 \\
        & AGILE3D~\cite{yue2023agile3d} & 24.8 & 45.7 & 72.4 \\
        & Ours  &    \textbf{39.2} & \textbf{64.6} & \textbf{85.5} \\
        \bottomrule
    \end{tabular}}
    \caption{Single-click evaluation for models trained only on ScanNet20, tested in the ScanNet20~$\rightarrow$~ScanNet20 (only seen objects) and ScanNet20~$\rightarrow$~ScanNet40 setting (seen + unseen objects).}
    \label{tab:non_interactive}
    \vspace{-0.3cm}
\end{table}

\vspace{0.1cm}
\subsection{Comparison with Non-Interactive Methods}
\label{res:non_interactive}
In Tab.~\ref{tab:non_interactive} we report a quantitative comparison with the non-interactive method Mask3D~\cite{schult2023mask3d} on seen and unseen ScanNet~\cite{dai2017scannet} classes, as reported originally in AGILE3D~\cite{yue2023agile3d}. This comparison follows the official \href{https://kaldir.vc.in.tum.de/scannet_benchmark/}{ScanNet Benchmark}, which evaluates methods by means of the Average Precision (AP). We follow the same evaluation protocol as AGILE3D~\cite{yue2023agile3d}\footnote{Numbers obtained by evaluating AGILE3D's official ScanNet40 pretrained model with its \href{https://github.com/ywyue/AGILE3D/tree/fcb8444ca2e2bd8270d349874b2e083c78e93b7d}{official implementation}.}, by taking a model trained only on ScanNet20 (20 object categories) and evaluating it on two datasets (referred as \textit{Setting} in Tab.~\ref{tab:non_interactive}): 1) 
 On ScanNet20 (ScanNet20~$\rightarrow$~ScanNet20), i.e. tested on the same 20 object categories \textit{seen} during training; 2) On ScanNet40 (ScanNet20~$\rightarrow$~ScanNet40), tested on the 20 \textit{seen} and on 20 \textit{unseen} object categories. Our model outperforms the other methods in both settings, with a clear margin on the ScanNet20~$\rightarrow$~ScanNet40 setting where methods are evaluated on \textit{seen+unseen} objects. As it will be seen in a targeted ablation study (Sec.~\ref{sec:ablation_click_fusion_neg_emb}), the difference in performance with respect to AGILE3D~\cite{yue2023agile3d} must be attributed to the click fusion, which allows the model to generalize better and achieve better performance even with a single click. 

\vspace{0.1cm}
\subsection{Ablation Studies}
\label{subsec:ablation}
To provide further insights of our method, we provide two ablation studies. They are targeted at testing its capabilities with different click fusion strategies and impact of the negative learned embeddings (Sec.~\ref{sec:ablation_click_fusion_neg_emb}), and to further test our model on unseen geometries represented by Gaussian Splatting~\cite{kerbl20233d} scenes (Sec.~\ref{sec:ablation_gaussian}). 

\vspace{-0.2cm}
\subsubsection{Click Fusion and Learned Negative Embedding}
\label{sec:ablation_click_fusion_neg_emb}
In this ablation study we want to highlight the impact of having an explicit vs. implicit click fusion, as well as how much the negative learned embeddings influence the segmentation performance. We provide the results in Tab.~\ref{tab:click_fusion}. To obtain these results, we trained our method only on ScanNet40 (following the main evaluation protocol as in Tab.~\ref{tab:interactive}) and evaluated it on ScanNet40~\cite{dai2017scannet}, S3DIS~\cite{armeni20163d} and KITTI-360~\cite{liao2022kitti} (\textit{Test Dataset}). We switched the click fusion strategy (\textit{Fusion}) between explicit and implicit (as described in Sec.~\ref{sec:fusion}), as well as switched on (\cmark) and off (\xmark) the use of the learned negative embedding (\textit{Neg.Emb.}).
With the results shown in Tab.~\ref{tab:click_fusion} we can claim that:
1) On unseen and challenging domains (i.e.~KITTI360), implicit fusion performs \textit{considerably better} than explicit fusion (\textbf{+73\%} IoU@10). On unknown datasets in fact, its likely that single-click masks have lower, more localized/spiked confidence, making explicit fusion \textit{slower} or even \textit{failing} to converge. Instead, with implicit fusion, the decoder reasons about all the clicks and labels \textit{together}, allowing it to still come up with a valid solution (see Fig.~4 and figures in the supplementary material);
2) In the same domain as training (i.e.~ScanNet40) and with many clicks, explicit fusion performs slightly better. The main reason behind this could be that with explicit fusion, the Nth click is likely to still have the highest confidence on the click's region, with \textit{consistent} impact on the final mask. Instead, with implicit fusion, all clicks are \textit{merged} and have \textit{proportionally decreasing} impact;
3) The negative embedding has a positive, consistent impact regardless of the click fusion.

\begin{table}
    \centering
    \resizebox{0.99\columnwidth}{!}{%
    \begin{tabular}{l|c|c|c|c|c|c|c}
        \toprule
        \textbf{Test Dataset} & \textbf{Fusion} & \textbf{Neg. Emb.} & \textbf{IoU@1} & \textbf{IoU@2} & \textbf{IoU@3} & \textbf{IoU@5} & \textbf{IoU@10} \\ 
        \toprule
        \multirow{4}{*}{ScanNet40} & \multirow{2}{*}{Explicit} &  \xmark & 59.6 & 68.0 & 73.2 & 78.0 & 82.6 \\ 
                                   &                           & \cmark & 62.7 & 70.5 & 75.2 & \textbf{79.6} & \textbf{83.6} \\  
        \cline{2-8}                
                                   & \multirow{2}{*}{Implicit} &  \xmark & 66.4 & 73.2 & 76.3 & 78.9 & 81.2 \\ 
                                   &                           & \cmark & \textbf{68.2} & \textbf{74.6} & \textbf{77.3} & \textbf{79.6} & 81.7 \\  
        \toprule
        \multirow{4}{*}{S3DIS}     & \multirow{2}{*}{Explicit} & \xmark & 53.8 & 67.0 & 73.2 & 81.4 & 85.9 \\  
                                   &                           & \cmark & 58.2 & 70.6 & 75.9 & 84.0 & \textbf{88.2} \\ 
        \cline{2-8}                
                                   & \multirow{2}{*}{Implicit} & \xmark & 63.4 & 73.9 & 79.5 & 84.0 & 87.2 \\  
                                   &                           & \cmark & \textbf{65.7} & \textbf{76.0} & \textbf{80.8} & \textbf{84.9} & 87.8 \\  
        \toprule
        \multirow{4}{*}{KITTI-360} & \multirow{2}{*}{Explicit} & \xmark & 31.0 & 37.8 & 40.0 & 43.6 & 46.3 \\ 
                                   &                           & \cmark & 34.5 & 41.0 & 42.6 & 45.9 & 48.2 \\ 
        \cline{2-8}                
                                   & \multirow{2}{*}{Implicit} & \xmark & 44.9 & 57.5 & 65.7 & 75.4 & 83.2 \\  
                                   &                           & \cmark & \textbf{46.3} & \textbf{58.7} & \textbf{66.7} & \textbf{76.2} & \textbf{83.6} \\    
        \bottomrule
    \end{tabular}}
    \caption{Ablation study highlighting the impact of the explicit vs. implicit click fusion, as well as the introduction of the learned negative embeddings into our model (trained only on ScanNet40).}
    \label{tab:click_fusion}
    % \vspace{-0.5cm}
\end{table}

\vspace{-0.2cm}
\begin{table}[]
    \centering
    \resizebox{0.9\columnwidth}{!}{%
    \begin{tabular}{l|l|l|c|c|c}
        \toprule
        \textbf{Training} & \textbf{Inference} & \multirow{2}{*}{\textbf{Method}} & \multirow{2}{*}{\textbf{IoU@1}} & \multirow{2}{*}{\textbf{IoU@2}} & \multirow{2}{*}{\textbf{IoU@3}} \\
        \textbf{Dataset} &  \textbf{Dataset}   &                  &              &                  &          \\
        \toprule
        \multirow{2}{*}{ScanNet40}                              & \multirow{2}{*}{GS-ScanNet40} & AGILE3D & 37.0 & 42.5 & 45.8 \\ 
                                      &                         & Ours & \textbf{44.9} & \textbf{55.4} & \textbf{61.5} \\
        \hline
        \multirow{2}{*}{GS-ScanNet40}                           & \multirow{2}{*}{GS-ScanNet40} & AGILE3D & 55.7 & 61.4 & 67.9 \\ 
                                      &                         & Ours & \textbf{63.9} & \textbf{71.3} & \textbf{74.8} \\
        \bottomrule
    \end{tabular}}
    \caption{Ablation study on the Gaussian Splatting domain.}
    \label{tab:gaussian}
    \vspace{-0.4cm}
\end{table}

\begin{figure*}
  \centering
  \begin{subfigure}{0.99\linewidth}
    \includegraphics[width=\textwidth]{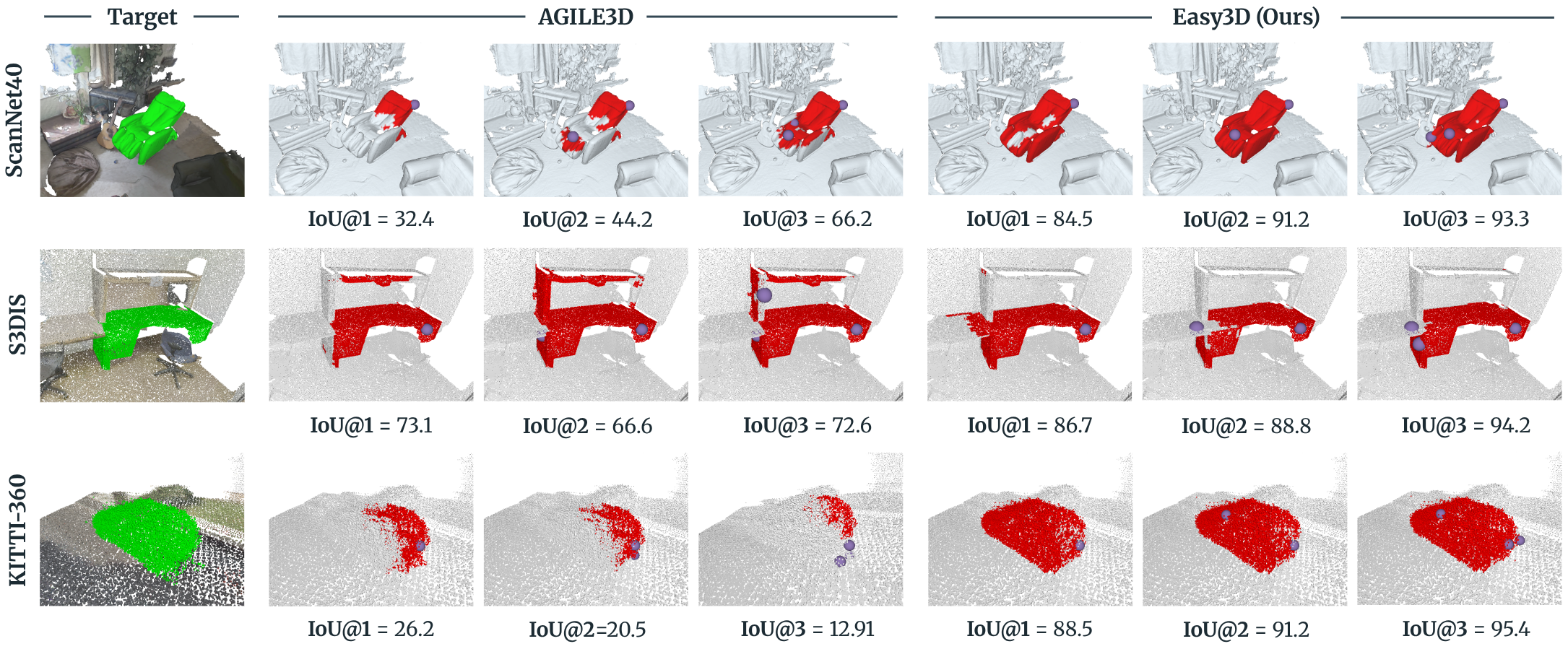}
  \end{subfigure}
  \caption{Given a target object (green mask, left) and a first click (sphere at IoU@1) we compare mask predictions (red) and IoU@i of our method and AGILE3D~\cite{yue2023agile3d} with $\mathsf{N_C}\leq3$. For 2nd and 3rd clicks we apply the same simulated interaction based on each method's errors.}
  \label{fig:qualcom}
  % \vspace{-0.15cm}
\end{figure*}

\begin{figure}
  \centering
  \begin{subfigure}{0.99\linewidth}
    \includegraphics[width=\textwidth]{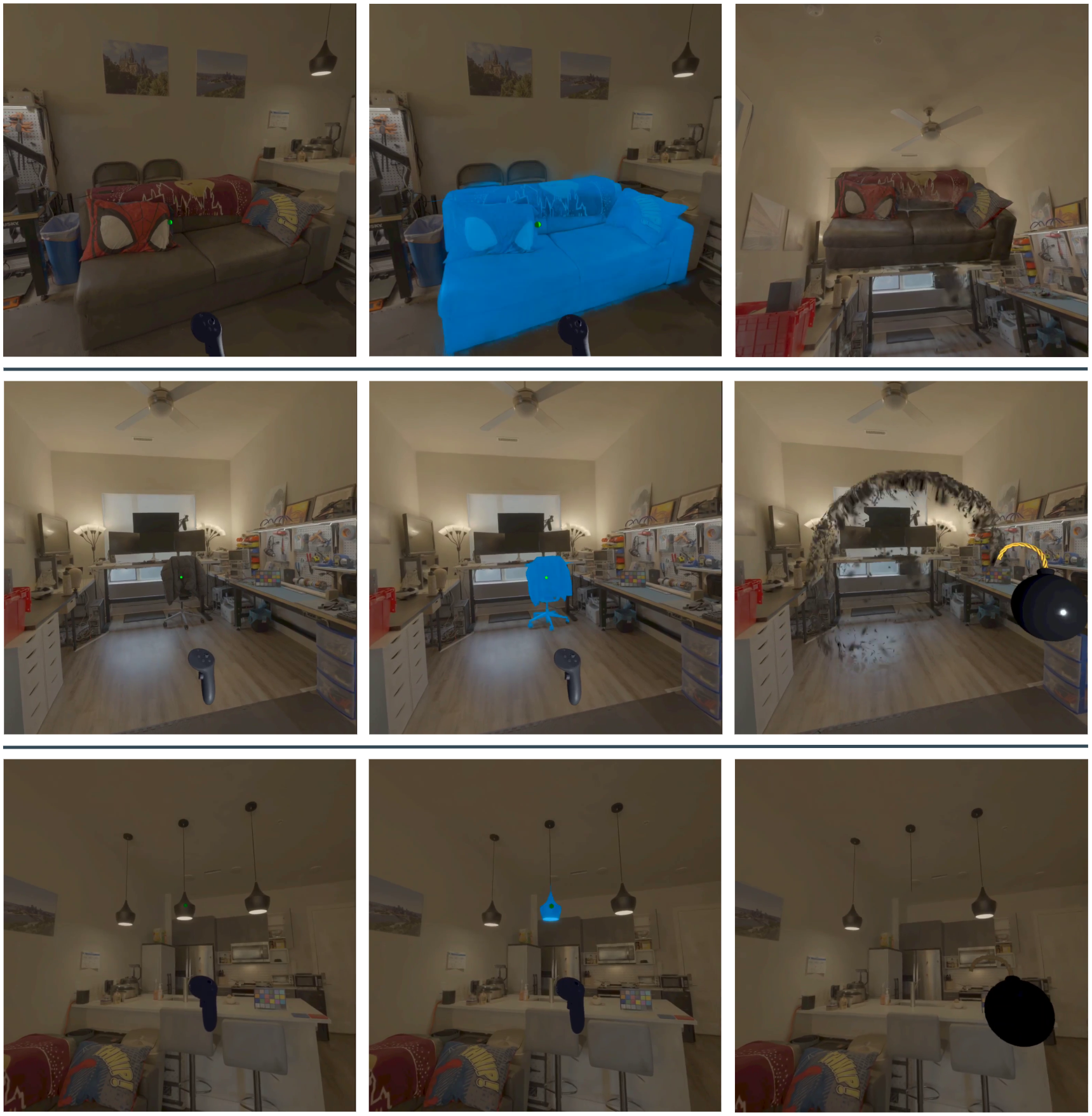}
  \end{subfigure}
  \caption{Visualizations of a VR application which, using a consumer-grade headset, lets a user easily segment objects using Easy3D, manipulate them (top row) and make them explode (middle and bottom rows). In this application, Easy3D fits seamlessly into a Gaussian Splatting~\cite{kerbl20233d} rendering pipeline enabling interactive 3D segmentation in real-time. The scene is taken from~\cite{xu2023vr}.}
  \label{fig:vr}
  \vspace{-0.40cm}
\end{figure}

\vspace{-0.10cm}
\subsubsection{Gaussian Splatting Scenes}
\label{sec:ablation_gaussian}
In Tab.~\ref{tab:gaussian} we ablate the segmentation results of our method and AGILE3D~\cite{yue2023agile3d} on a Gaussian Splatting (GS)~\cite{kerbl20233d} version of ScanNet40 (GS-ScanNet40). The GS-ScanNet40 dataset has been obtained with the popular \href{https://docs.nerf.studio/nerfology/methods/splat.html}{SplatFacto} repository~\cite{nerfstudio}, and 3D Instance Segmentation labels have been ported from ScanNet40~\cite{dai2017scannet}. More details about GS-ScanNet40 will be provided in the supplementary material. Tab.~\ref{tab:gaussian} shows that our method, in both Mesh~$\rightarrow$~GS and GS~$\rightarrow$~GS settings, clearly outperforms AGILE3D~\cite{yue2023agile3d}. AGILE3D~\cite{yue2023agile3d} results for GS~$\rightarrow$~GS have been obtained by training it on GS-ScanNet40 using its \href{https://github.com/ywyue/AGILE3D/tree/fcb8444ca2e2bd8270d349874b2e083c78e93b7d}{official repository}. 

\section{Qualitative Results}
\label{sec:qualitative_results}

In Fig.~\ref{fig:qualcom} we visualize the predictions of our method and AGILE3D~\cite{yue2023agile3d} for up to 3 user clicks, reporting the corresponding IoU@\{1,2,3\} metric between the predicted mask (red mask) and the ground-truth object (green mask), of models trained on ScanNet40~\cite{dai2017scannet}. Our method shows superior generalization performance, especially on the most challenging domain represented by KITTI-360 (bottom row). In Fig.~\ref{fig:vr} we provide snapshots from a VR application featuring Easy3D, where a user can define 3D clicks to segment objects (blue masks) move them and re-arrange them (top row) as well as make them \textit{explode} (middle, bottom row). We provide this as an example of real application for our method, which in this case fits seamlessly in a state-of-the-art Gaussian Splatting~\cite{kerbl20233d} rendering pipeline.
% \vspace{-0.10cm}
\section{Conclusions}
We proposed Easy3D, an Interactive 3D Instance Segmentation method which features an architecture that maximizes simplicity, performance and generalization. The method relies on a voxel-based encoder, on a transformer-based decoder and on implicit click fusion. We also show the first application of the negative embedding in implicit fusion, which is demonstrated to help the network improve its generalization capabilities. Our method achieves consistent state-of-the-art performance on both seen and unseen domains (\ie~ScanNet~\cite{dai2017scannet}, S3DIS~\cite{armeni20163d}, KITTI-360~\cite{liao2022kitti}, ScanNet++~\cite{yeshwanth2023scannet++}), as well on a newly-introduced Gaussian Splatting~\cite{kerbl20233d} representation of ScanNet40, particularly on unseen objects and challenging settings, demonstrating its suitability for a wide variety of applications.

\newpage

{
    \small
    \bibliographystyle{ieeenat_fullname}
    \bibliography{main}
}

\clearpage
\setcounter{page}{1}
\maketitlesupplementary

% \cref{sec:intro}

\noindent This supplementary material provides:
\begin{itemize}
\item Additional details on the decoder architecture and attention operations (Sec.~\ref{sec:decoder});
\item Details on ScanNet++ and the creation of the Gaussian Splatting version of ScanNet40 (GS-ScanNet40) (Sec.~\ref{sec:gs});
\item Additional implementation details of our method (Sec.~\ref{sec:impl_details}); 
\item Additional qualitative results (Sec.~\ref{sec:qual}).  
\end{itemize}

\section{Additional Decoder details}
\label{sec:decoder}

\begin{figure}
  \centering
  \begin{subfigure}{0.99\linewidth}
    \includegraphics[width=\textwidth]{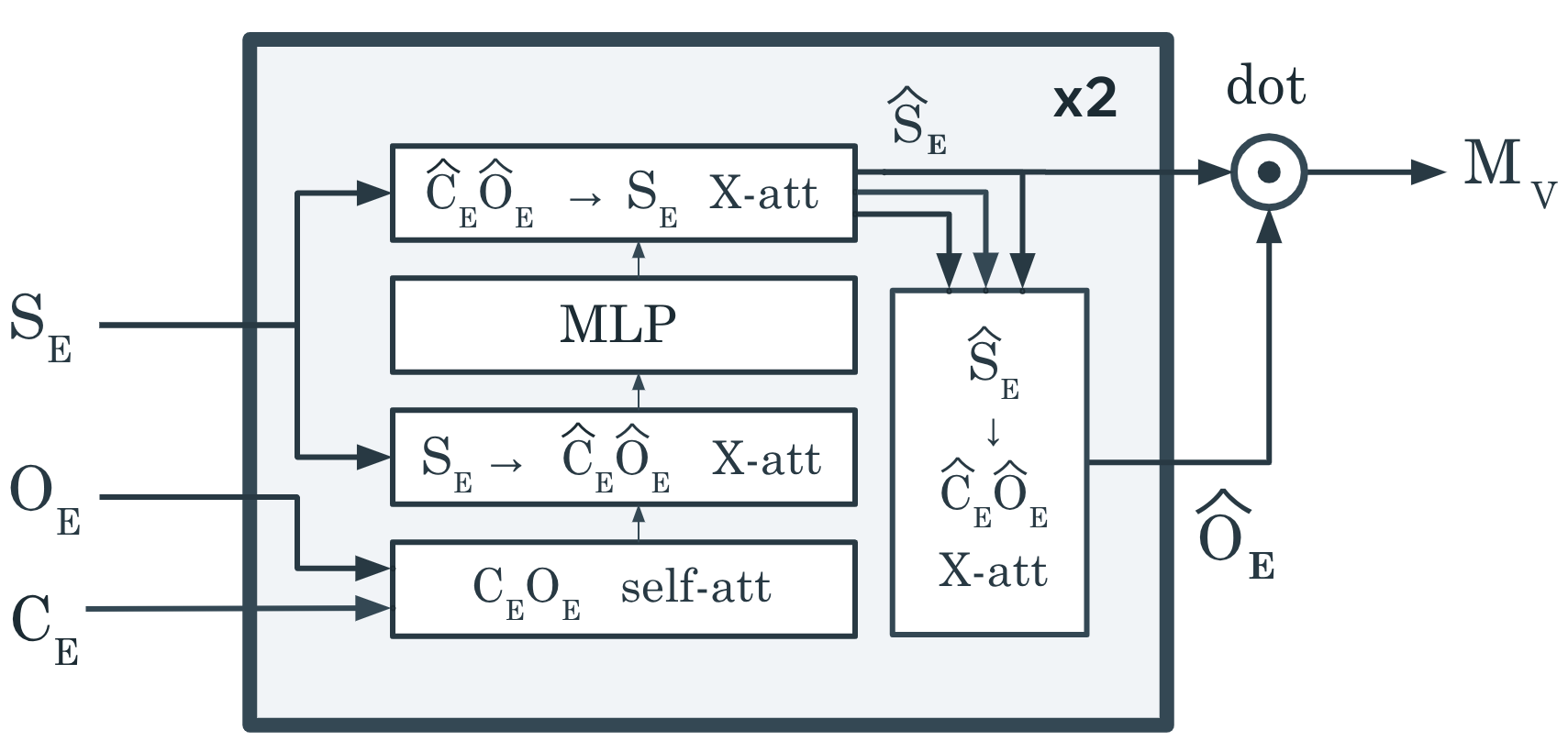}
  \end{subfigure}
  \caption{Detail of the attention operations performed inside our decoder.}
  \label{fig:sam_decoder}
  \vspace{-0.40cm}
\end{figure}

\begin{figure}
  \centering
  \begin{subfigure}{0.5\linewidth}
    \includegraphics[width=\textwidth]{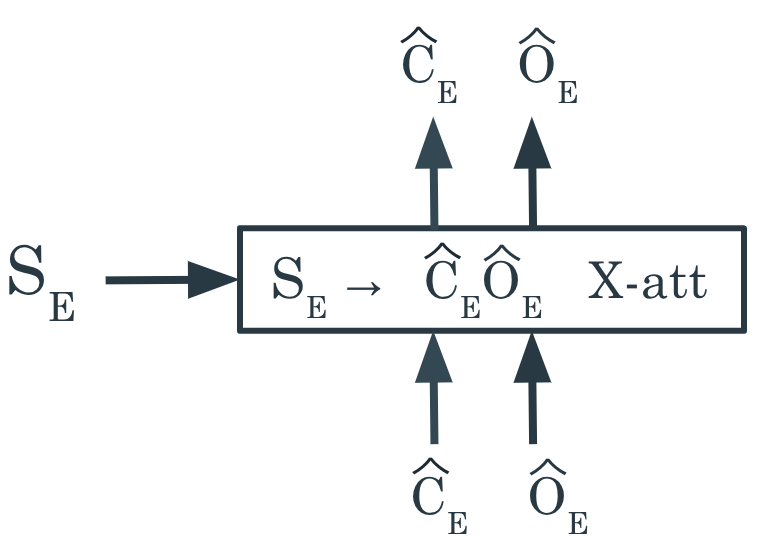}
  \end{subfigure}
  \caption{Detail of one of the attention-based~\cite{vaswani2017attention} operations performed in our decoder.}
  \label{fig:attn}
  \vspace{-0.40cm}
\end{figure}

In our decoder, depicted in Fig.~\ref{fig:sam_decoder}, we use a combination of attention-based ~\cite{vaswani2017attention} layers which are used to exchange information between the scene $S_E$, clicks $C_E$ and output $O_E$ embeddings. We use the symbol "$\rightarrow$" to describe the \textit{direction of the exchange of information} that a specific layer is performing. More in detail, we refer to the definition of attention operation in \cite{vaswani2017attention} which relies on \textit{queries}, \textit{keys} and \textit{values}. In the operation in Fig.~\ref{fig:attn}, which is part of our decoder, the concatenation of clicks and output embedding ($\widehat{C_E} \widehat{O_E}$) will correspond to the queries, while the keys and values will be represented by the scene embedding ($S_E$). Note that, as done in SAM~\cite{kirillov2023segment}, in any attention operation we add to the queries and keys their corresponding positional encoding and label encoding if needed. In case an attention operation involves the use of clicks, we add their corresponding positional encoding and learned label encoding each time, while in case it involves the scene embedding, we add the voxels' positional encoding. 
\vspace{-0.2cm}

\section{Details on GS-ScanNet40 and ScanNet++}
\label{sec:gs}
In order to test the capabilities of our method on a different geometric distribution, we introduced a Gaussian Splatting~\cite{kerbl20233d} version of ScanNet40, i.e.~GS-ScanNet40. To create it, we relied on the \textit{SplatFacto} method available in the popular \href{https://github.com/nerfstudio-project/nerfstudio}{Nerfstudio} repository~\cite{nerfstudio}. We reconstructed all the $\approx$ 1500 ScanNet40 scenes using their corresponding posed DSLR images, following the standard SplatFacto configuration. Once obtained the GS reconstruction, we matched the ScanNet40 mesh-based instance annotations with their gaussian version, by finding the nearest point to the mesh for each gaussian, and assigning the corresponding instance label if their distance was below 5cm. Note that 5cm is the voxel resolution $V_S$ used in all our experiments. Please also note that, being ScanNet a dataset where 3D annotations have been performed and are available only on incomplete and decimated meshes, it was not possible to use images to encode the labels into the gaussians during the scene reconstruction. This process in fact requires the definition of many heuristics, which eventually lead to an unsatisfactory solution. 

For what concerns ScanNet++~\cite{yeshwanth2023scannet++}, we used the \href{https://github.com/scannetpp/scannetpp}{official repository} to obtain vertex-level instance annotations on the meshes. We then obtained our results on the 50 scenes that are part of the official validation set, retaining the objects that are part of the official Instance Segmentation \href{https://kaldir.vc.in.tum.de/scannetpp/benchmark/insseg}{benchmark} at the time of submission. 
\vspace{-0.125cm}

\section{Additional implementation details}
\label{sec:impl_details}
We trained our method using a batch size of 8 on an NVIDIA A100 with 80GB of memory for $\approx$ 2 days, using an AdamW~\cite{loshchilovdecoupled} optimizer with a 0.05 weight decay.
\vspace{-0.125cm}

\section{Additional Qualitative Results}
\label{sec:qual}
\begin{figure*}
    \centering
    \vspace{0.5cm}
    \includegraphics[width=\textwidth]{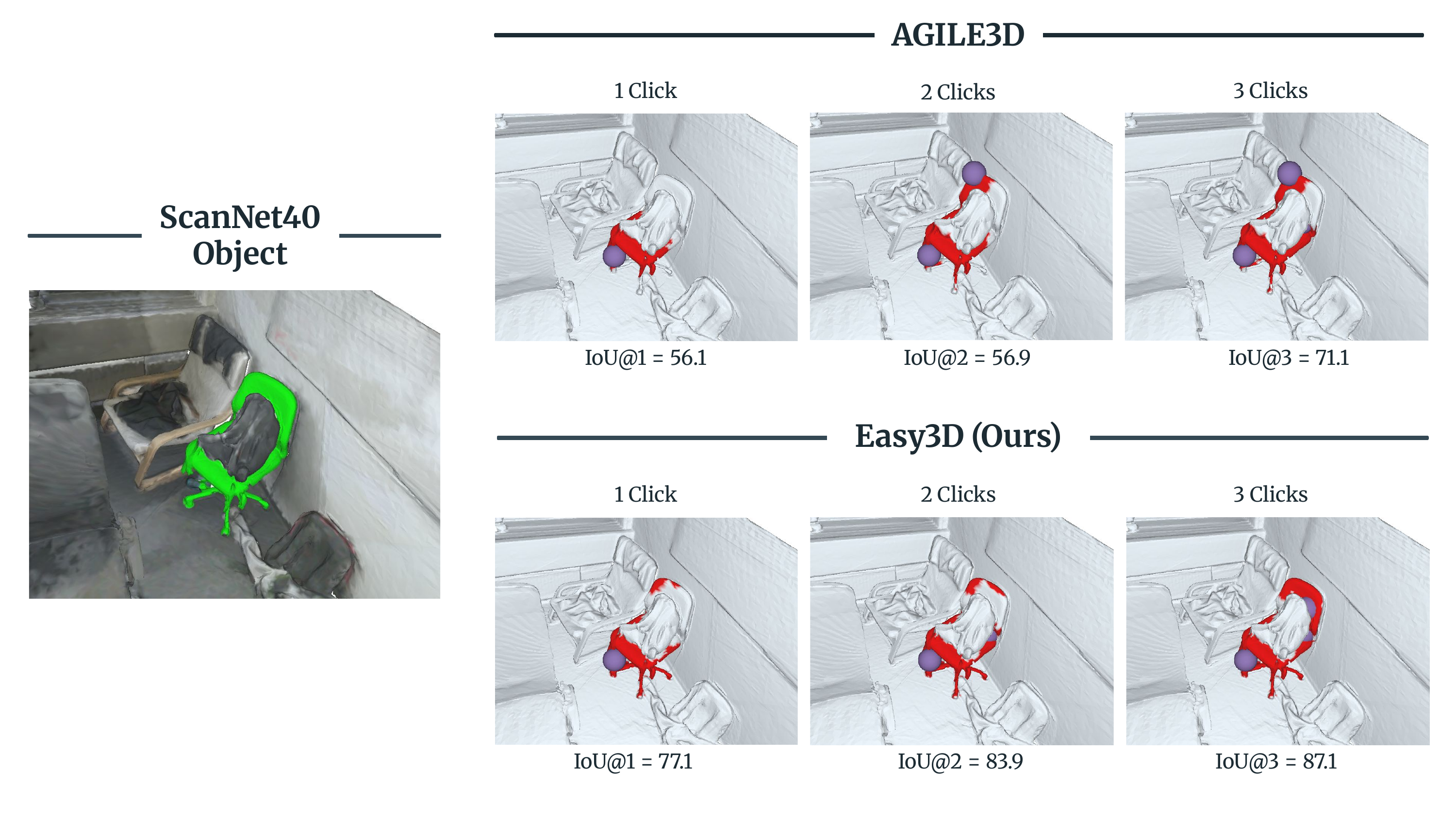} \\
    \vspace{1.0cm}
    \includegraphics[width=\textwidth]{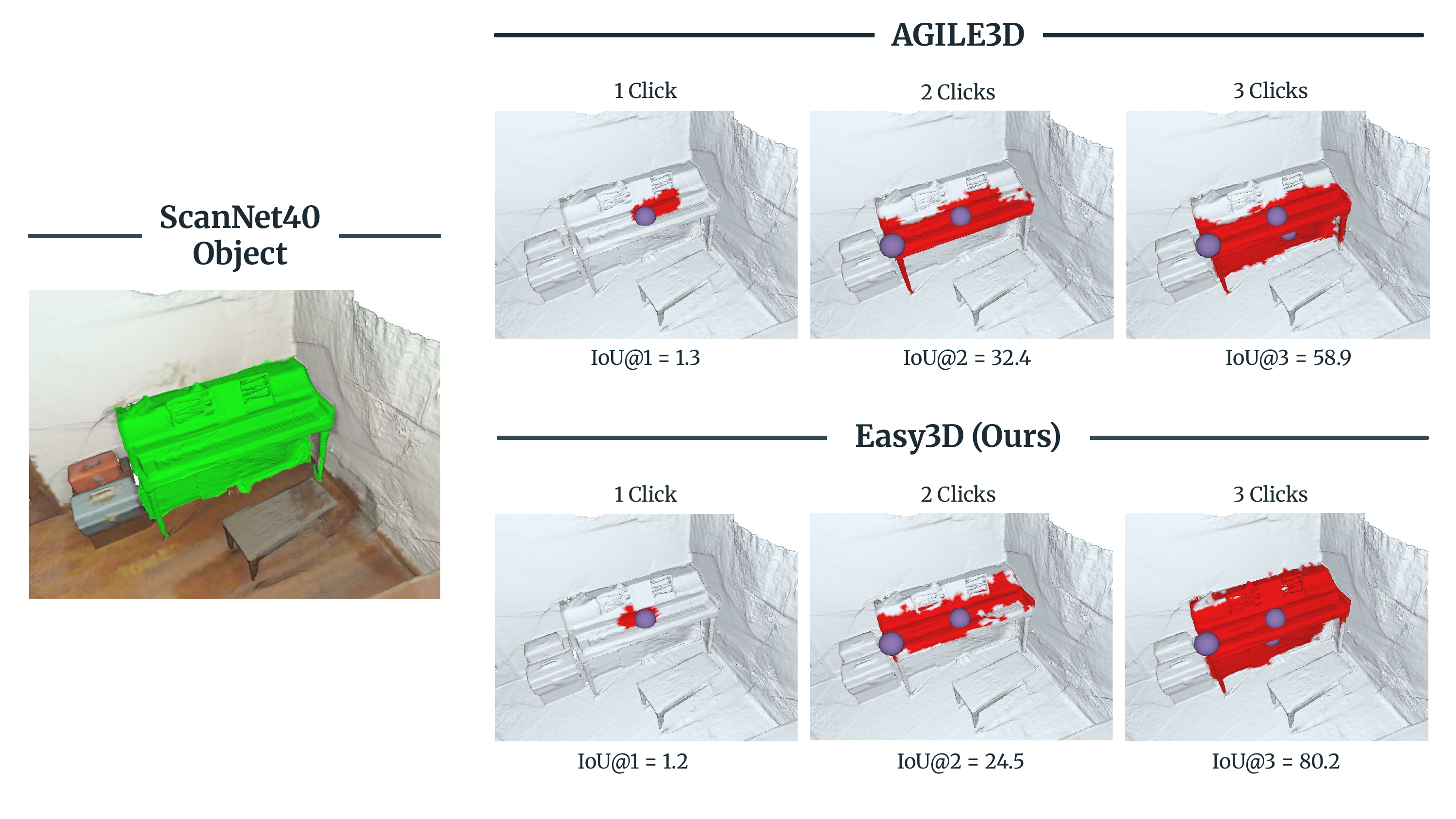}
    \caption{Additional qualitative results on ScanNet40~\cite{dai2017scannet}.}
    \label{fig:qualsupp1}
\end{figure*}

\begin{figure*}
    \centering
    \vspace{0.5cm}
    \includegraphics[width=\textwidth]{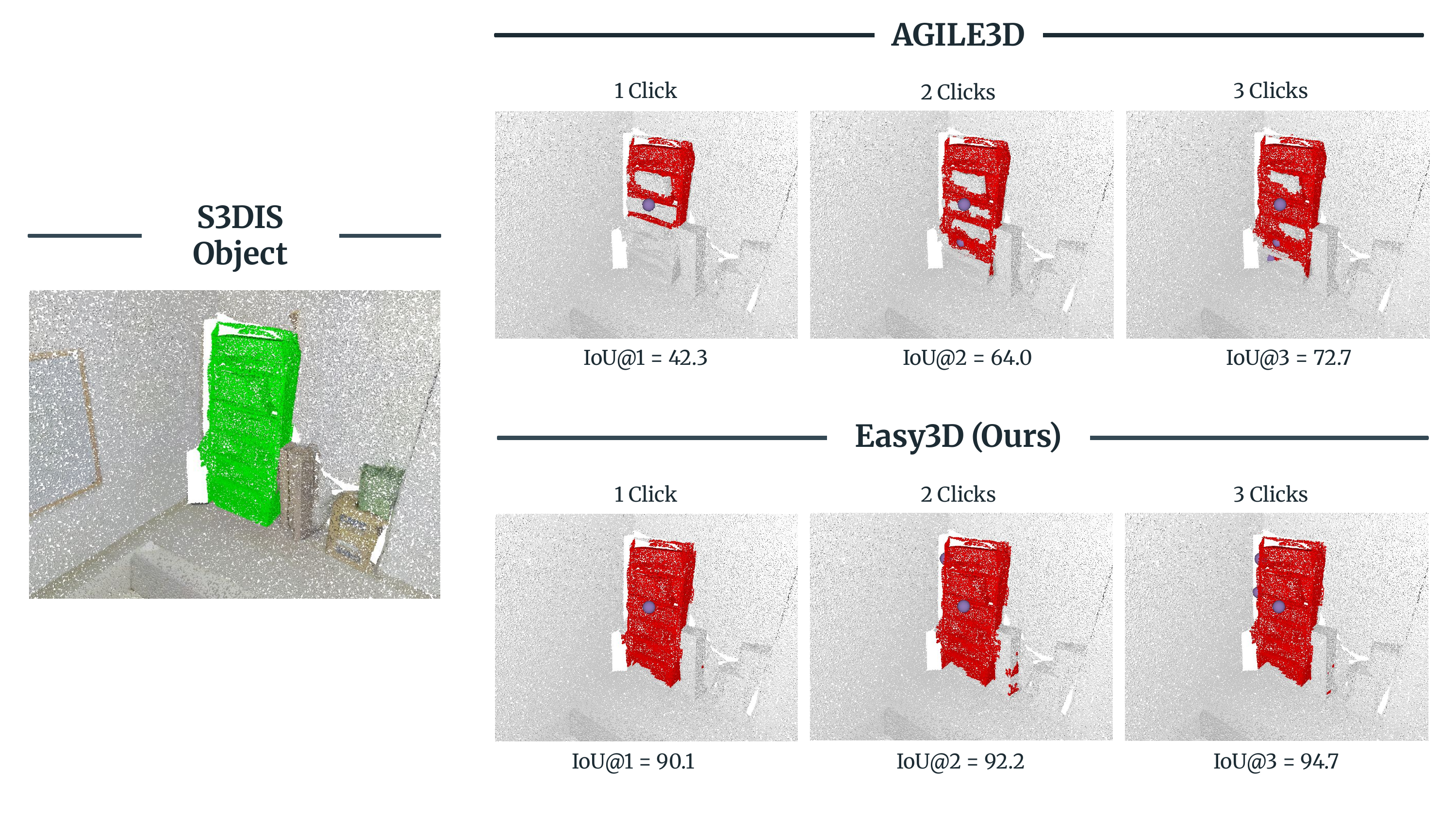} \\
    \vspace{1.0cm}
    \includegraphics[width=\textwidth]{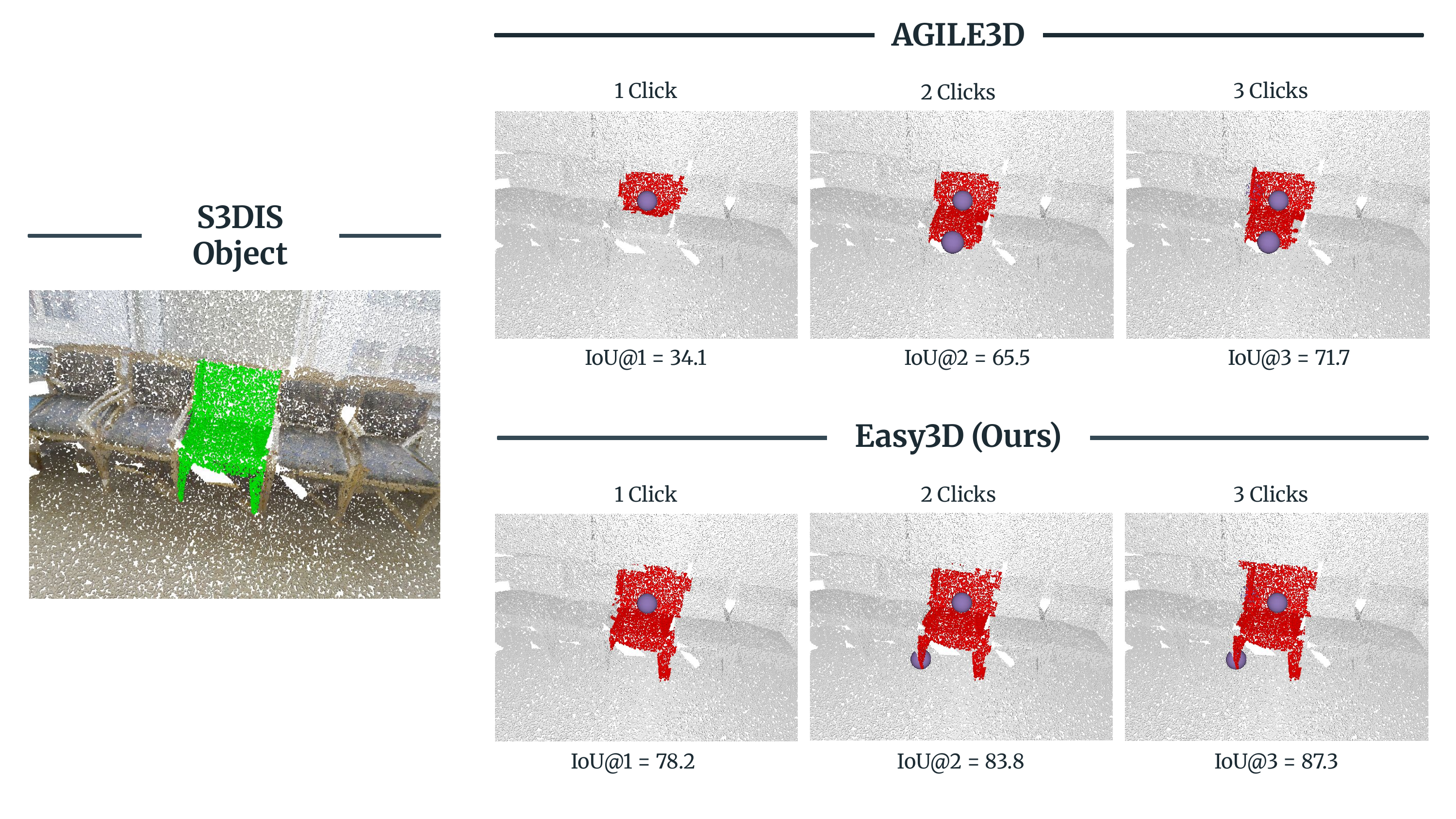}
    \caption{Additional qualitative results on S3DIS~\cite{armeni20163d}.}
    \label{fig:qualsupp2}
\end{figure*}

\begin{figure*}
    \centering
    \vspace{0.5cm}
    \includegraphics[width=\textwidth]{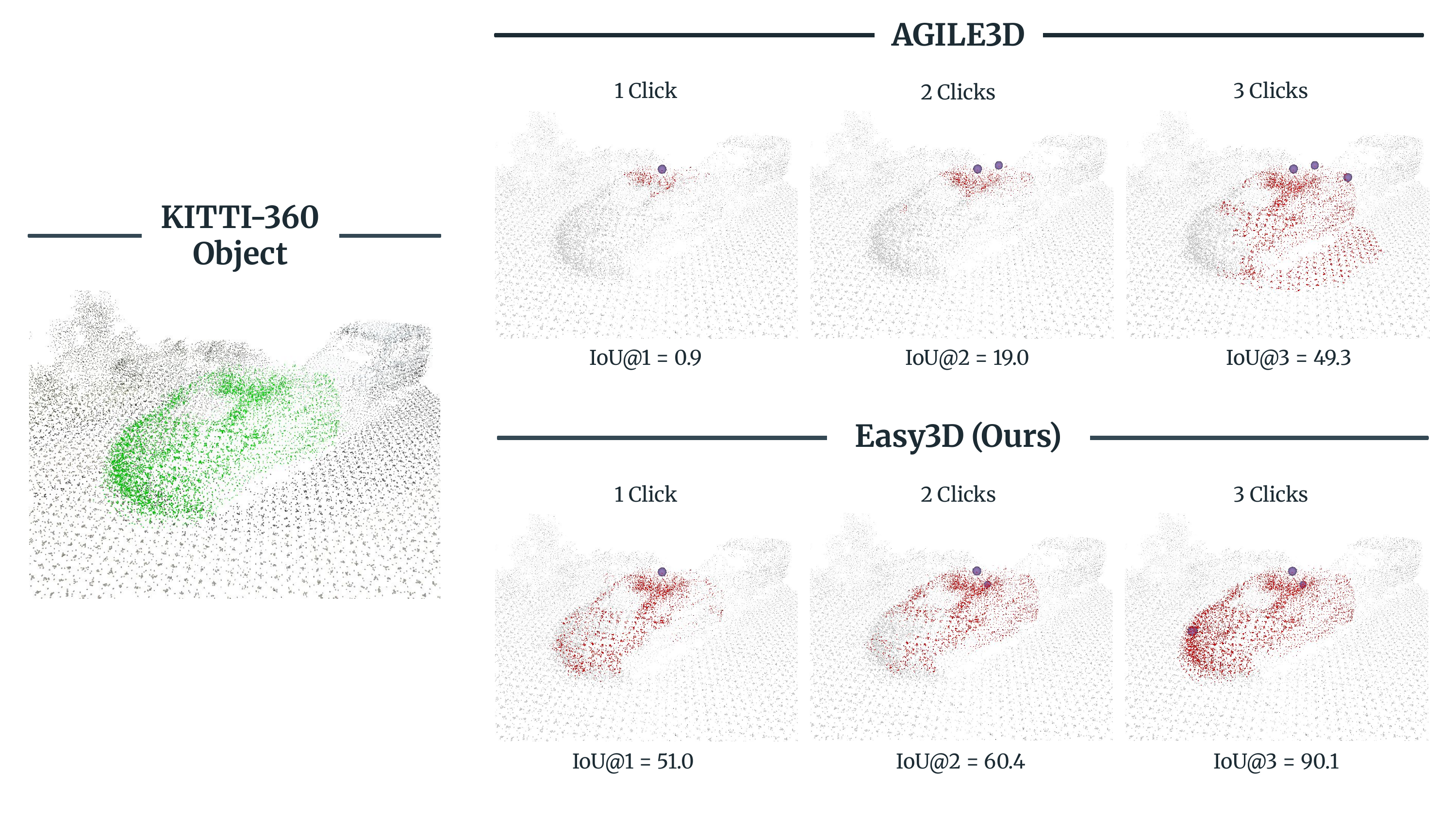} \\
    \vspace{1.0cm}
    \includegraphics[width=\textwidth]{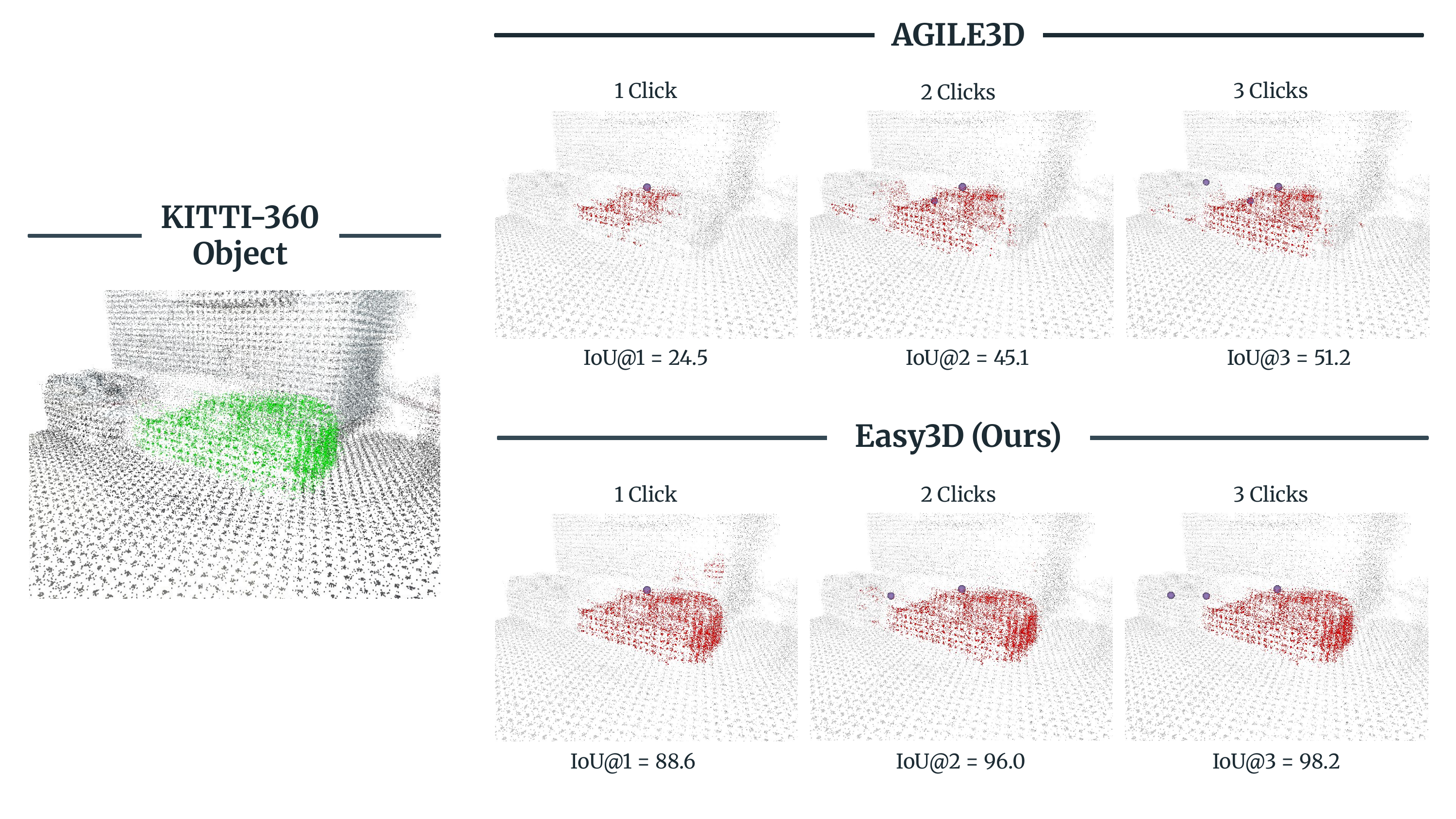}
    \caption{Additional qualitative results on KITTI-360~\cite{liao2022kitti}.}
    \label{fig:qualsupp3}
\end{figure*}

\begin{figure*}
    \centering
    \vspace{0.5cm}
    \includegraphics[width=\textwidth]{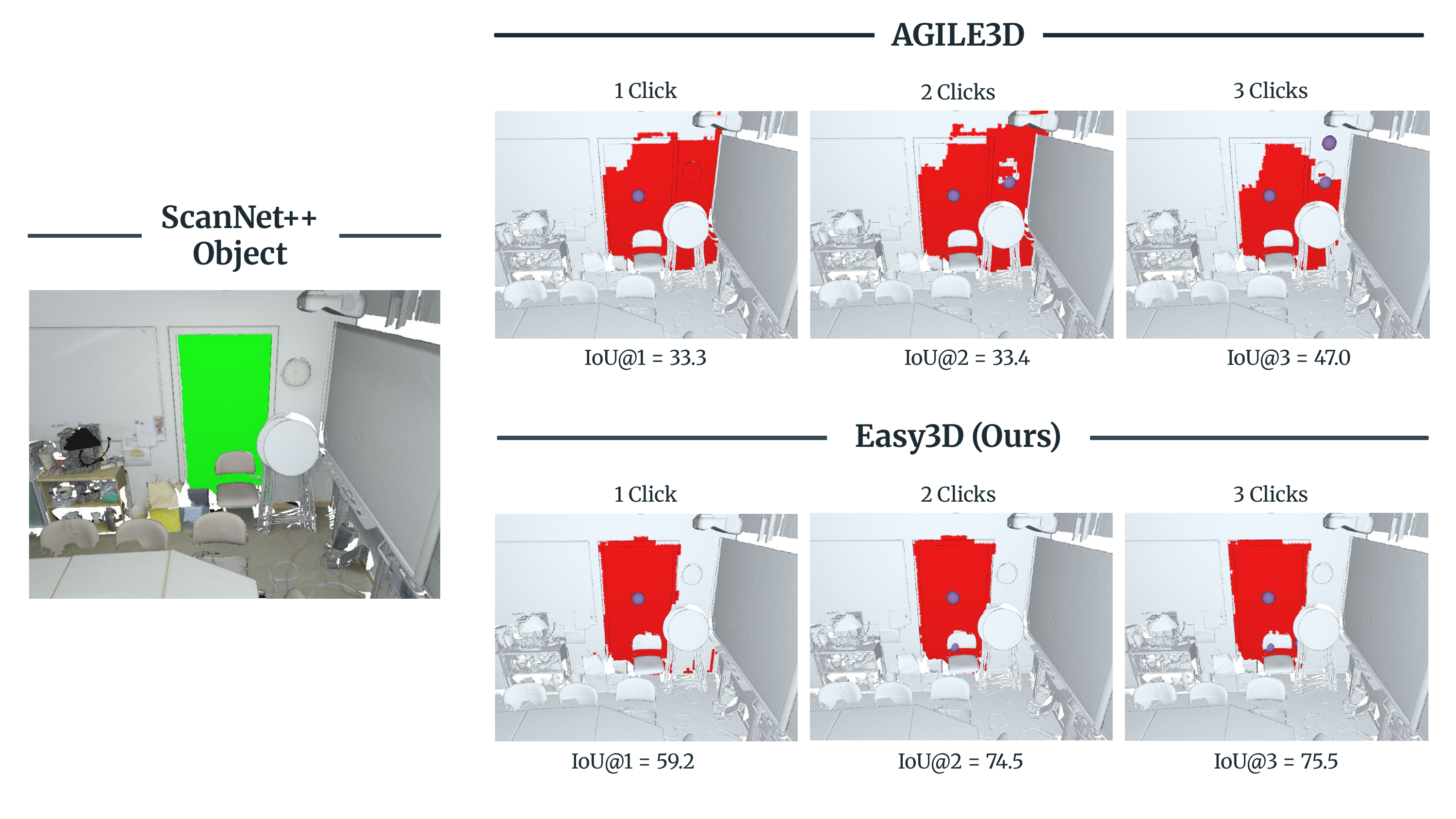} \\
    \vspace{1.0cm}
    \includegraphics[width=\textwidth]{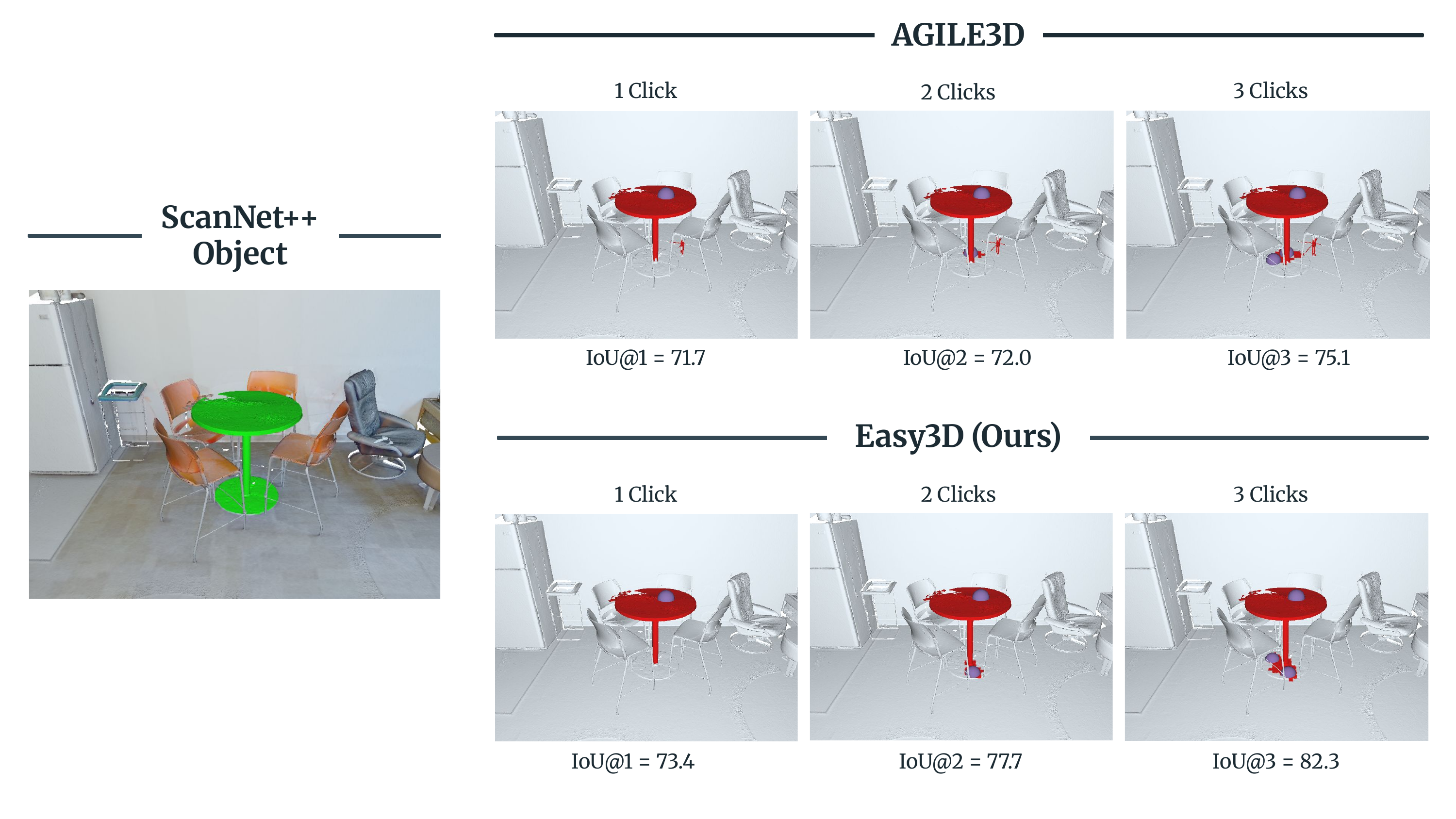}
    \caption{Additional qualitative results on ScanNet++~\cite{yeshwanth2023scannet++}.}
    \label{fig:qualsupp4}
\end{figure*}

In \cref{fig:qualsupp1,fig:qualsupp2,fig:qualsupp3,fig:qualsupp4} we provide additional qualitative results by visualizing the predictions of our method and AGILE3D~\cite{yue2023agile3d}. Similarly to the main paper, we visualize results for up to 3 user clicks reporting the corresponding IoU@\{1,2,3\} metric between the predicted mask (red mask) and the ground-truth object (green mask) of models trained only on ScanNet40~\cite{dai2017scannet}. Please note that while the clicks have been visualized with a similar blue sphere, they can represent a positive or negative click depending if they are part of the ground-truth object mask shown on the left (green). If a click is on the mask, then the click will be positive, otherwise it will be negative. 

\end{document}